\pdfoutput=1 
\def\year{2021}\relax
\documentclass[letterpaper]{article} 
\usepackage{aaai21}  
\usepackage{times}  
\usepackage{helvet} 
\usepackage{courier}  
\usepackage[hyphens]{url}  
\usepackage{graphicx} 
\def\UrlFont{\rm}  
\usepackage{natbib}  
\usepackage{caption} 
\usepackage{grffile}

\usepackage{amsfonts,amsthm,amsmath}
\usepackage{braket}
\usepackage{balance}
\usepackage[ruled,linesnumbered]{algorithm2e} 
\usepackage{booktabs}

\usepackage{xspace} 
\usepackage{multirow}

\newcommand{\dataSet}{\textsl}
\newcommand{\themepark}{\dataSet{ThemePark}\xspace}
\newcommand{\routecity}{\dataSet{RouteCity}\xspace}

\newcommand{\methodFont}{\textsl}
\newcommand{\random}{\methodFont{Random Walk}\xspace}
\newcommand{\markov}{\methodFont{Markov Model}\xspace}
\newcommand{\rnn}{\methodFont{RNN Model}\xspace}
\newcommand{\gan}{\methodFont{GAN Model}\xspace}

\newcommand{\ours}{\methodFont{MoveSD}\xspace}
\newcommand{\ourswo}{\methodFont{\ours wo/ dynamic}\xspace}
\newcommand{\ourswogail}{\methodFont{\ours wo/ \gail}\xspace}
\newcommand{\gail}{\methodFont{GAIL}\xspace}

\newcommand{\Policy}{\pi}
\newcommand{\System}{\mathcal{G}}
\newcommand{\judge}{\mathcal{J}}
\newcommand{\Prediction}{f}

\newtheorem{definition}{Definition}
\newtheorem{problem}{Problem}

\newcommand{\goal}{g}

\newcommand{\problemSetFont}{\mathcal}
\newcommand{\traj}{\problemSetFont{\tau}}
\newcommand{\Traj}{\problemSetFont{T}}
\newcommand{\TrajD}{\problemSetFont{T}^D}
\newcommand{\TrajG}{\problemSetFont{T}^G}

\newcommand{\trajS}{\problemSetFont{\xi}}
\newcommand{\TrajS}{\problemSetFont{S}}

\newcommand{\Expect}{\mathbb{E}}
\newcommand{\Real}{\mathbb{R}}
\newcommand{\Loss}{\mathcal{L}}
\newcommand{\Disc}{\mathcal{D}}
\newcommand{\DiscT}{\mathcal{D}_\psi}

\newcommand{\ie}{\mbox{\it{i.e.}}}
\newcommand{\eg}{\mbox{\it{e.g.}}}

\title{How Do We Move: Modeling Human Movement with System Dynamics}
\author{
Hua Wei,
Dongkuan Xu,
Junjie Liang,
Zhenhui Li \\
}

\affiliations {
    College of Information Sciences and Technology, The Pennsylvania State University \\
    \{hzw77, dux19, jul672, jessieli\}@ist.psu.edu
}

\begin{document}
\maketitle

\begin{abstract}
Modeling how human moves in the space is useful for policy-making in transportation, public safety, and public health. The human movements can be viewed as a dynamic process that human transits between states (\eg, locations) over time. In the human world where intelligent agents like humans or vehicles with human drivers play an important role, the states of agents mostly describe human activities, and the state transition is influenced by both the human decisions and physical constraints from the real-world system (\eg, agents need to spend time to move over a certain distance). Therefore, the modeling of state transition should include the modeling of the agent's decision process and the physical system dynamics. 
In this paper, we propose \ours to model state transition in human movement from a novel perspective, by learning the decision model and integrating the system dynamics. \ours learns the human movement with Generative Adversarial Imitation Learning and integrates the stochastic constraints from system dynamics in the learning process. To the best of our knowledge, we are the first to learn to model the state transition of moving agents with system dynamics. In extensive experiments on real-world datasets, we demonstrate that the proposed method can generate trajectories similar to real-world ones, and outperform the state-of-the-art methods in predicting the next location and generating long-term future trajectories.
\end{abstract}

\section{Introduction}

Modeling how human moves in space is useful for policy-making in various applications, ranging from transportation~\cite{wu2018stabilizing,lian2014geomf}, public safety~\cite{wang2017no} to public health~\cite{wang2018inferring}. For example, modeling the movements of vehicles with human drivers can help build a good simulator and serve as a foundation and a testbed for reinforcement learning (RL) on traffic signal control~\cite{wei2018intellilight} and autonomous driving~\cite{wu2018stabilizing}. In fact, lacking a good simulator is considered as one of the key challenges that hinder the application of RL in real-world systems~\cite{dulac2019challenges}. 

\begin{figure}[t!]
  \centering
  \includegraphics[width=0.45\textwidth]{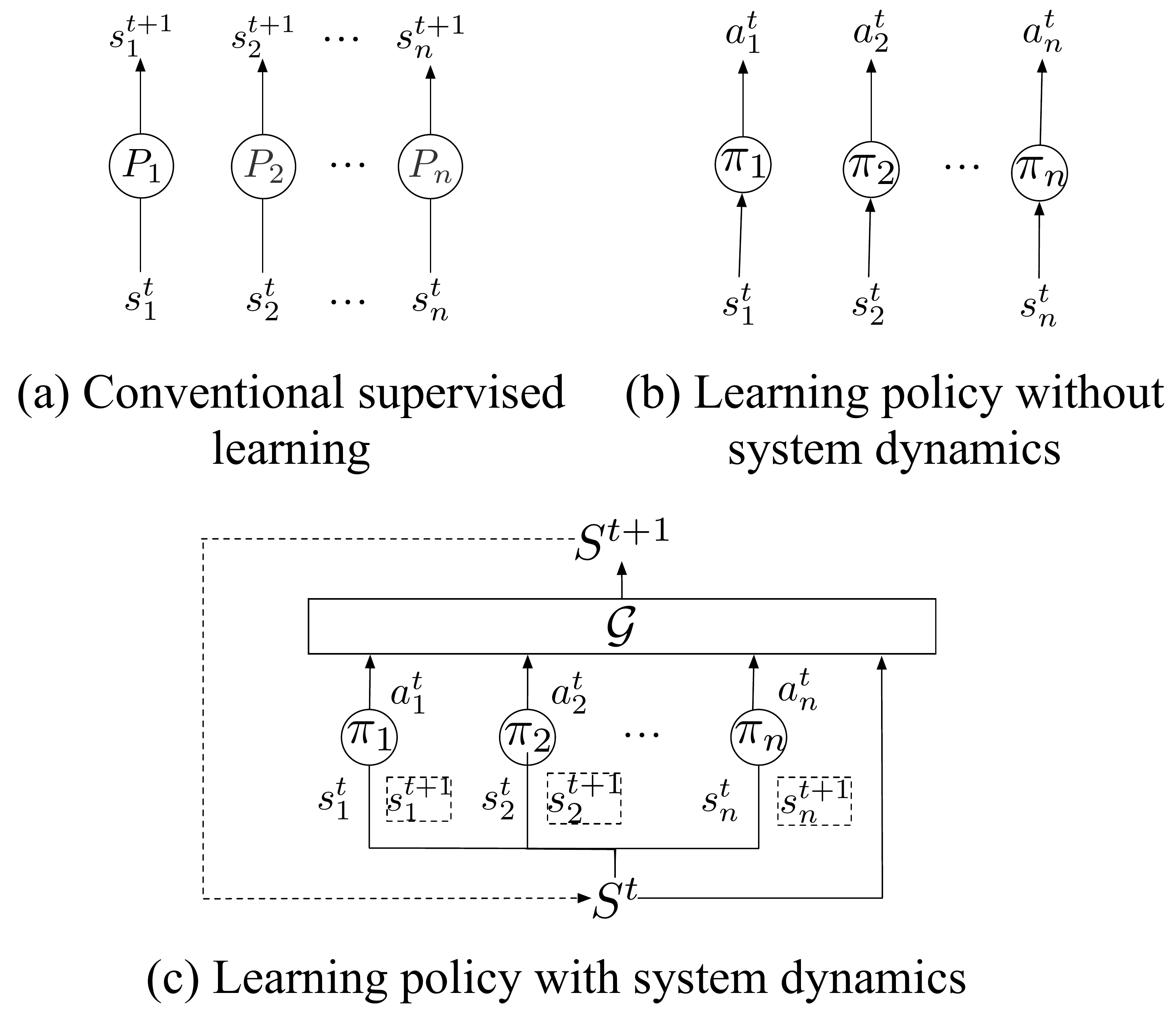}
  \caption{Three perspectives on modeling state transitions. (a) Traditional methods directly predict the next state $s^{t+1}$ for an agent based on its past states with a model $P$.
  (b) Direct learning of agent policy $\Policy$ that maps from state $s$ to action $a$.
  (c) Learning agent policy $\Policy$ with the system dynamics $\System$. The agents observe its state $s^t$ from the system state $S^t$, which is modeled by system dynamics model $\System$.}
  \label{fig:intro}
  \vspace{-3mm}
\end{figure}

The modeling of human movements in space can be viewed as modeling the state transition of intelligent agents (\eg, human travelers, or vehicles with human drivers). The state of an agent can include its historical locations and its destination. 
Recently there are growing research studies that have developed computational methods to model the state transitions. 
One line of research directly predict the next states based on the current state and historical states~\cite{song2016prediction,liu2016predicting,feng2017poi2vec,baumann2018selecting,feng2018deepmove,gupta2018social,sadeghian2019sophie,ma2019trafficpredict,finn2016unsupervised,oh2015action}. As is shown in Figure~\ref{fig:intro} (a), these methods focus on directly minimizing the error between estimated state $\hat{s}_{t+1}$ and true observed state $s_{t+1}$, with an end-to-end prediction model $P$ for the agent.
In the real world where agents can take actions and interact with each other, the state of an agent is influenced not only by its previous states, but more importantly, by its action and the actions of other agents. It would require a large number of observed states that cover the whole state distribution to learn an accurate state transition model. When the agent encounters unobserved states, the learned model is likely to predict $s^{t+1}$ with a large error.
An example is, when a vehicle drives on the road it has never observed before, these methods would fail to predict the movements. If we know the vehicle's driving policy (\eg, the vehicle follows the shortest path to its destination), the vehicle's movement can still be inferred.

Another line of research considers the underlying mechanism behind the state transition of agent movements from a decision-making perspective~\cite{ziebart2008maximum,zou2018understanding,bhattacharyya2018multi,song2018multi}. As shown in Figure~\ref{fig:intro} (b), they aim to learn a decision policy $\Policy$, which can generate the movements by taking a sequence of actions $a$ from policy model $\pi$. For example, imitation learning (IL) can be used to learn the routing policy of a vehicle, which aims to learn to take actions (\eg, keep moving on the current road, turn left, turn right, go straight, and take U-turn) based on the current state of the vehicle (\eg, the road ID, the number of vehicles and average speed on the road)~\cite{ziebart2008maximum}.
The learned policy is more applicable to the unobserved state because the dimension of action space is usually smaller than state space that increases exponentially with the number of state features. 
However, IL methods assume that the next state is purely decided by the action of the agent, while in the real world, the state transition of an agent is a combined effect of both agent decision and system dynamics. 
For example, if a driver presses the brake pedal, system dynamics determine this vehicle's location after this action, which is mostly affected by factors such as the tires of vehicles, road surface, and weather.
As another example, a person arrives at location $A$ and wants to check in, but $A$ has a limitation in the population it can serve. So the person would have to spend time waiting. Therefore, if we ignore the fact that the policy and final state of an agent must comply with the constraints from system dynamics, it will make the learned state transition model less realistic.

With the limitations of traditional prediction methods and imitation learning methods, in this paper, we formulate the problem of state transition modeling as modeling a decision-making process and incorporating system dynamics. Figure~\ref{fig:intro} (c) illustrates our formulation of the problem: in a system with $N$ agents, we consider the state of an agent as a joint outcome of its decision and system dynamics. At each timestep $t$, the agent observes state $s^t$ from the system state $S^t$, takes action $a^t$ following policy $\Policy$ at every time step. Then the system model $\System$ considers current system state $S^t$ and the actions $\{a_1^t,\cdots, a_N^t \}$ of all agents and outputs the next system state $S^{t+1}$, upon which the agents observe their next states $\{s_1^{t+1},\cdots, s_N^{t+1} \}$. This formulation looks into the mechanism behind state transitions and provides possibilities to include system dynamics in the modeling of state transitions.

In this paper, we propose \ours, which utilizes a similar framework as Generative Adversarial Imitation Learning (\gail)~\cite{ho2016generative} to model agents' decision process, with a generator learning movement policy $\Policy$ and a discriminator $\Disc$ learning to differentiate the generated movements from observed true movements. Moreover, \ours explicitly models system dynamics $\System$ and its influence on the state transition through learning $\Policy$ and $\Disc$ with the constraints from $\System$ and through providing an additional intrinsic reward to $\Policy$. Extensive experiments on real-world data demonstrate that our method can accurately predict the next state of an agent and accurately generate longer-term future states. In summary, our main contributions are:
~\noindent\\$\bullet$ We present the first attempt to learn to model the state transition of moving agents with system dynamics. Specifically, we formulate the state transition of human movements from the decision-making perspective and learn the movement policy under the framework of \gail.
~\noindent\\$\bullet$ We show the necessity to consider the stochastic dynamics of the system when modeling state transitions and provide insights on different possible approaches to integrate the system dynamics.
~\noindent\\$\bullet$ We perform extensive experiments on real-world data, and the experimental results show that our method can generate similar human movement trajectories to the true trajectories, and has superior performances in applications like location prediction and trajectory generation compared with the state-of-the-art methods.


\section{Preliminaries}
In this section, we formulate our problem of modeling the state transition of human movements and then illustrate our definition using an example of a traveler moving in the grid world.


\begin{figure*}[ht]
  \centering
  \includegraphics[width=.9\linewidth]{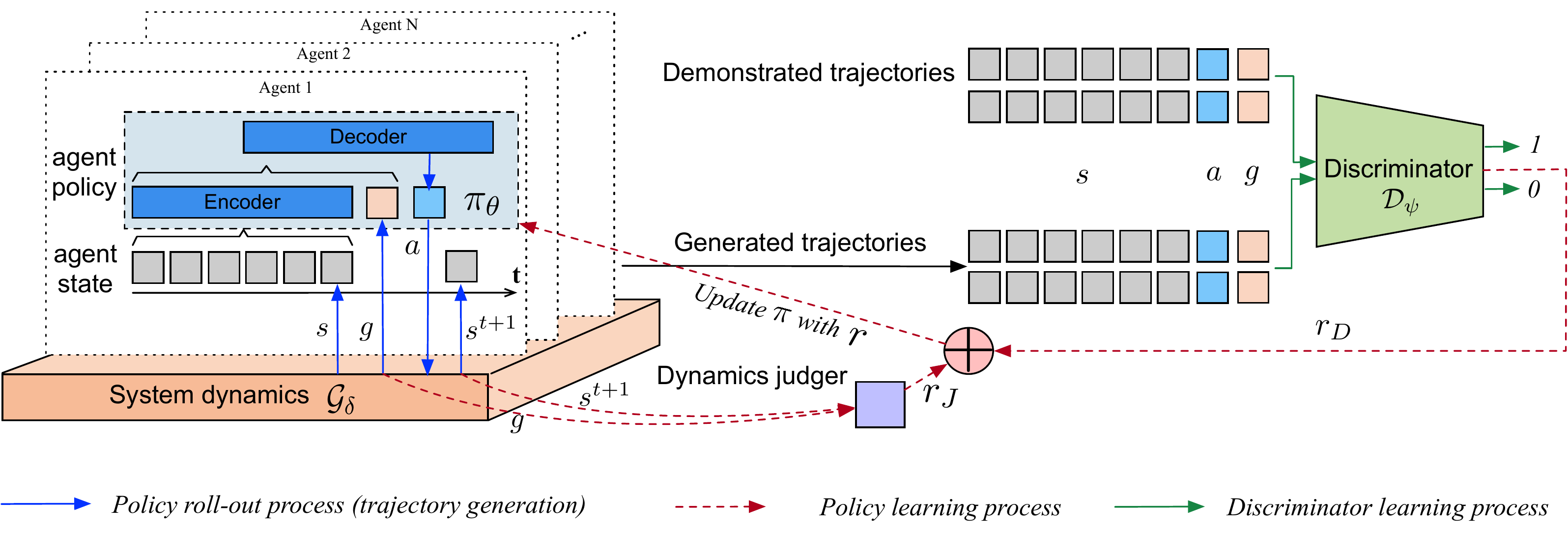}
  \vspace{-3mm}
  \caption{Illustration of the proposed framework. It has four main components: the system dynamics $\System_\delta$, the dynamic judger $J$, the agent policy $\Policy_\theta$, and the discriminator $\Disc_\psi$. $\Policy$ and $\System$ together influences the transition of states. The discriminator learns to differentiate between demonstrated trajectories and generated trajectories and provide reward $r_D$ for learning policy $\Policy$ that behaves similarly to the true policy. The dynamics judger provides additional intrinsic reward $r_J$ from the system to the policy, indicating whether the policy behaves compliantly with system constraints $\goal$. Better viewed in color.}
  \label{fig:framework}
  \vspace{-3mm}
\end{figure*}

\begin{definition}[State and action]
A state $s^t$ of an agent describes the surrounding environment of the agent at time $t$, and the action $a^t$ is the agent takes at time $t$ using its policy $\Policy$, i.e., $a^t\sim\Policy(a|s^t)$.
\end{definition}

\begin{definition}[State trajectory and movement trajectory]
A state trajectory of an agent is a sequence of states generated by the agent, usually represented by a series of chronologically ordered pairs, i.e., $\trajS = ( s^{t_0},\cdots ,s^{t_T} )$. A movement trajectory of an agent is a chronologically ordered sequence of state-action pairs, i.e., $\traj = (\traj^{t_0},\cdots ,\traj^{t_T} )$ where $\traj^{t_i}=(s^{t_i}, a^{t_i})$ .
\end{definition}

\begin{problem}
Given a set of state trajectories $\TrajS = \{ \trajS_1, \trajS_2,\cdots, \trajS_N \}$ of a real-world system with $N$ agents, the overall objective is to learn the transition function $f(s^{t+1}|s^{t})$ from $s^t$ to $s^{t+1}$ so that the error between estimated state $\hat{s}^{t+1}$ and true state $s^{t+1}$ is minimized.
\end{problem}

As in Figure~\ref{fig:intro} (a), traditional supervised learning solutions to this problem learn a direct mapping function $\Prediction_\vartheta$ from $s^t$ to $s^{t+1}$ through minimizing some loss function $\Loss$ over the set of state trajectories as training data: $
    \mathop{argmin}_{\vartheta} \Expect_{(s^t,s^{t+1})\sim \Traj_E}[\Loss(s^{t+1},\Prediction_\vartheta(s^t)]$.

In this paper, instead of learning a direct mapping from $s^t$ to $s^{t+1}$, we tackle the problem from a decision-making perspective. As shown in Figure~\ref{fig:intro} (c), in our problem, an agent observes state $s^t$ from the system state $S^t$, takes action $a$ following policy $\Policy$ at every time step. System model $\System$ takes current system states $S^t$ and the actions $\{a_1^t,\cdots, a_N^t \}$ as input, and decides the next system state $S^{t+1}$, upon which the agents observe their states $\{s_1^{t+1},\cdots, s_N^{t+1} \}$.

To better understand our definition, we use an example of learning to model traveler movements in a grid world, where the travelers move between grids, and there are several facilities scattered in the grids that the travelers might be interested.
~\noindent\\$\bullet$ \textbf{State}. In this example, the state $s$ of an agent includes the traveler's information, \eg, coordinates of the current grid the traveler locates. State observations could also include the environment properties, \eg, the population in the current grid as well as events.
~\noindent\\$\bullet$ \textbf{Action}. The traveler's action is its moving direction, which can mostly be inferred from its two consecutive locations. For example, if a traveler is at grid $A$ at time $t$ and its next location is $A$'s southwest neighboring grid, its action $a^t$ is ``move southwest''. 
~\noindent\\$\bullet$ \textbf{Agent}. Each traveler is an agent. A traveler takes action $a$ based on the current state $s^t$ according to his policy $\Policy$, \eg, a traveler decides to move to the next location, keep wandering in the current location, or check-in at the facilities in current location based on his current state. The policy could vary for different agents, \eg, different travelers have different movement policies.
~\noindent\\$\bullet$ \textbf{System dynamics}. System dynamics $\System$ aggregates the current actions of all agents $\{s_1^t,\cdots, s_N^t \}$ in the system, and influencing their next states $\{s_1^{t+1},\cdots, s_N^{t+1} \}$. For example, if a location has an event and has a queue in serving travelers, system dynamics determine the constraint $\goal_i$ like how long an upcoming traveler $i$ would wait to get served and influence the traveler's next state. Note that $\System$ takes the system state $S$ as input, whereas the state $s$ of an agent is only part of the system state $S$ observed by the agent. The system dynamics model $\System$ is different from the agent policy $\Policy$, where $\System$ reflects the physical transitions and constraints of the real world. In contrast, $\Policy$ is the mapping from the agent's observation to action. 
~\noindent\\$\bullet$ \textbf{Objective}. This example's objective is to minimize the difference between the observed traveler movement trajectories and the traveler trajectories generated by the learned transition model $f_{\theta, \delta}=\{ \Policy_\theta, \System_\delta \}$, where $\Policy$ and $\System$ are parameterized by $\theta$ and $\delta$.

\section{Method}
In this section, we first overview the general framework of \ours; then, we describe each component of the architectures; finally, we introduce the training process and discuss possible paradigms in learning the models.

\subsection{Overview}
As is shown in Figure~\ref{fig:framework}, \ours has four components, namely the system dynamics $\System_\delta$, the dynamic judger $J$, the agent policy $\Policy_\theta$, and the discriminator $\Disc_\psi$ to model the state transition. We formulate the problem of learning the transition model $f_{\theta, \delta}=\{ \Policy_\theta, \System_\delta \}$ to perform real-world-like movements by rewarding it for ``deceiving'' the discriminator $\Disc_\psi$ trained to discriminate between policy-generated and observed true trajectories.  

The system dynamics $\System_\delta$ takes as input the system state $S$ and generates the physical constraints $\goal$ for each agent, which we parameterize as a multilayer perceptron (MLP). The policy $\Policy_\theta$ takes as input the observed trajectories and the physical constraints $\goal^t$ from $\System_\delta$ and generate an action distribution $\Policy_\theta(a|s)$ and sample an action $\hat{a}^t$ from the distribution, which we parameterize as a Recurrent Neural Network (RNN): $\hat{a}^t \sim \pi(a | s^{t-L},\cdots,s^{t}; \goal^t)$,
where $L$ is the observed trajectory length.

The next state $s^{t+1}$ of the agent can be calculated based on $s^{t}$, $\hat{a}^{t}$ and with system states $S^t$. Then the movement trajectories $\TrajG=\{ \traj_1,\cdots, \traj_N \}$ of $N$ agents in the system can be generated. The generated trajectories $\TrajG$, are then fed to 1) the discriminator $\DiscT$ to output a score of the probability of the trajectory being true and 2) the dynamics judger $\judge$,  a rule-based calculator, to measure the extent of state $s$ meeting system physical constraints $\goal$. 

The policy $\Policy_\theta$ and discriminator $\DiscT$ are jointly optimized through the framework of \gail ~\cite{ho2016generative} in the form of an adversarial minimax game as Generative Adversarial Network (GAN)~\cite{goodfellow2014generative}:
\begin{equation}
\label{eq:gail}
\begin{aligned}
    \mathop{max}_{\psi} \mathop{min}_{\theta} \Loss(\psi,\theta) = \Expect_{(s,a;\goal)\sim \traj \in \Traj_E} \log\DiscT(s,a;\goal) + \\ \Expect_{(s,a;\goal)\sim \traj \in \Traj_G} \log (1-\DiscT(s,a;\goal)) - \beta H(\Policy_\theta)
\end{aligned}
\end{equation}
where $\Traj_E$ and $\Traj_G$ are the observed true trajectories and the trajectories generated by $\Policy_\theta$ and $\System_\delta$ in the system, $H(\Policy_\theta)$ is an entropy regularization term. Different from the vanilla \gail whose discriminator is trained with $(s, a)$, in this paper, the discriminator is also conditioned on the system constraints as $\DiscT(s,a;\goal)$.

\subsection{Stochastic System Dynamics}
\label{sec:method:dynamic}
System dynamics $\System$ can influence the movements of the agents in the form of system constraints $\goal$, including the temporal and the spatial constraints. \emph{The temporal constraint}, i.e., the time for an agent to spend in a location, is particularly crucial in deciding the action of the agent. For example, due to the physical distance between two locations and the agent's travel speed, it is unlikely for an agent to move arbitrarily from one location to another instantly. Another example is that the duration a traveler stays at a demanding location is influenced by the waiting time to get served. \emph{The spatial constraint}, like obstacles that block the agent's actions, is usually more stable than temporal constraints and can be learned with features like location ID. Therefore, in this work, we consider temporal constraint as the main influence of system dynamics underlying the agent decision-making process. 

However, the stochasticity of system dynamics poses challenges to learn the system constraints since some factors in the system might be intrinsically unobserved and the system dynamics are naturally stochastic. For example, the duration of vehicles traveling on a specified road is dynamically changing with weather conditions or traffic signals, where we do not always observe the weather and traffic signal situation in the agent's trajectory data. 

To learn the stochastic constraints, instead of directly predicting a scalar value of the temporal constraint like traditional supervised learning methods do~\cite{finn2016unsupervised, oh2015action,song2010limits, song2016prediction,feng2018deepmove,liu2016predicting}, we set out to learn the latent distribution of the temporal constraint.
Since the temporal constraint $\goal$, \ie, the duration an agent stays in the location, is a real-valued scalar with lower and upper bounds (correspondingly 0 and maximum simulation steps in our cases), we can shift and rescale the values to be in the range $[0, 1]$ and model $\goal$ as a sample from a Beta distribution $Beta(\Xi)$, where $\Xi = (\alpha, \beta)$ is the 2-dimensional shaping parameter for Beta distribution $(\alpha, \beta > 0)$. Specifically, $\System$ is parameterized with an MLP:
\begin{equation}
    \begin{aligned}
    h^g_0 = & \sigma(o_g W^g_0+b^g_0), h^g_1 = \sigma(h^g_{0}W^g_1+b^g_1), \cdots  \\
    \Xi =  & \sigma(h^g_{G-1}W^g_G+b^g_G), \goal \sim Beta(\Xi)
    \end{aligned}
\end{equation}
where $G$ is the number of layers, $W^g_i\in\Real^{n_{i-1} \times n_{i}}$, $b^g_i\in\Real^{n_i}$ are the learnable weights for the $i$-th layer. $\sigma$ is ReLU function (same denotation for the following part of this paper), as suggested by~\cite{radford2015unsupervised}. For $i=0$, we have $W^g_0\in\Real^{c \times n_0}$, $b^g_0\in\Real^{n_0}$, where $c$ is the feature dimension for $o_g$ and $n_0$ is the output dimension for the first layer; for $i=G$, we have $W^g_G\in\Real^{n_{G-1} \times 2}$, $b^g_G\in\Real^{1\times 2}$. 

\subsection{Policy Network}
The policy network consists of three major components: 1) observation embedding; 2) recurrent encoding; 3) action prediction. 

 
\subsubsection{Observation Embedding}

We embed $k$-dimensional state features into an $m$-dimensional latent space via an embedding layer that copes with location ID $loc^t$, a layer of MLP for the rest of the state features $o^t$ and a concatenation layer on the outputs from previous two layers:
\begin{equation}
\begin{aligned}
x^t 
    & = Concat(OneHot(loc^t)W_e, \sigma(o^t W_o+b_o))
\end{aligned}
\end{equation}
where $W_e\in\Real^{l\times m_1}$, $W_o\in\Real^{k\times m_2}$ , $b_o\in\Real^{m_2}$ are weight matrixes and bias vectors to learn. Here, there are total $l$ location IDs, and $m_2=m-m_1$. The concatenated state represents the current state of the agent.

\subsubsection{Recurrent encoding}
To reason about the status of the agent as well as the dynamic interactions between multiple agents, unlike conventional feed-forward neural policies~\cite{ho2016generative,li2017inferring}, we propose to learn the policy $\Policy$ with an encoder-decoder model~\cite{cho2014learning} to account for the sequential nature. 
To capture the individual status from its past observations, we input the observations of past $L_{in}$ timesteps to the encoder RNN, one observation per step, which progresses as: $h_R^i = RNN_{enc}(x^i, h^{i-1}), \forall t-L_{in}\leq i\leq t$.
The last hidden state, $h_R^t$ is the fixed-length descriptor of past trajectories for the agent. 

\subsubsection{Action prediction}
The action prediction module takes the output of encoder RNN $h_R^t$, and the system dynamic constraints $\goal^t$ as input. The final action is sampled from the categorical distribution $p^A$ learned by an MLP:
\begin{equation}
    \begin{aligned}
    h^A_0 & = Concat(h_R^t, \goal^t),\\
    h^A_1 & = \sigma(h^A_0 W^A_1+b^A_1),\ h^A_2 = \sigma(h^A_{1}W^A_2+b^A_2), \cdots  \\
   p^A & =  Softmax(h^A_{H-1}W^A_H+b^A_H), a^t \sim Cat(p^A)
    \end{aligned}
\end{equation}
where $W^A_i\in\Real^{d_{i-1} \times d_{i}}$, $b^A_i\in\Real^{d_i}$ are the learnable weights for the $i$-th layer in action prediction module. For $i=H$, we have $W^A_H\in\Real^{n_{H-1} \times |\mathcal{A}|}$, $b^A_H\in\Real^{|\mathcal{A}|}$, where $|\mathcal{A}|$ is the total number of candidate actions.

\subsection{Discriminator and Dynamics Judger}
The discriminator network takes a similar network structure as a policy network, with the action prediction module in $\Policy$ replaced by a binary classifier with an MLP.
When training $\DiscT$, Equation~\eqref{eq:gail} can be set as a sigmoid cross entropy where positive samples are from observed true trajectories $\Traj_E$, and negative samples are from generated trajectories $\Traj_G$. Then optimizing $\psi$ can be easily done with gradient ascent on the following loss function:

\begin{equation}
\label{eq:loss}
    \centering
    \begin{aligned}
    \Loss_D & = \Expect_{(s,a;\goal) \in \Traj_E} \log\DiscT(s,a;\goal) \\ & +  \Expect_{(s,a;\goal) \in \Traj_G} \log (1-\DiscT(s,a;\goal)) 
    \end{aligned}
\end{equation}

The transition between states in a real-world system is an integration of physical rules, control policies, and randomness, which means its true parameterization is assumed to be unknown. Therefore, given $\Traj_G$ generated by $\Policy_\theta$ in the system, Equation~\eqref{eq:gail} is non-differentiable w.r.t $\theta$ and the gradient cannot be directly back-propagated from $\DiscT$ to $\Policy_\theta$. 
Therefore, we learn $\Policy_\theta$ through Trust Region Policy Optimization (TRPO)~\cite{Schulman2015Trust} in reinforcement learning, with a surrogate reward function formulated from Equation~\eqref{eq:gail} as:
\begin{equation}
\label{eq:gail-reward}
    r_D(s^t, a^t; \goal^t) = - \log(1 - \DiscT(s^t, a^t; \goal^t))
\end{equation}
Here, the surrogate reward $r_D(s^t, a^t, \goal^t)$ is derived from the discriminator $\DiscT$ and can be perceived to be useful in driving $\Policy_\theta$ into regions of the state-action space at time $t$ similar to those in observations.

\paragraph{Dynamics judger $\judge$} To avoid generating trajectories that do not meet the real-world constraints, it is necessary to provide guidance from system dynamics during the learning of policy $\Policy_\theta$.
To explicitly incorporate the system constraints, we draw inspiration from~\cite{ding2019goal} and propose a novel intrinsic reward term into the learning of our policy as follows:
\begin{equation}
\label{eq:gail-reward1}
    r_J(s^t, a^t,\goal^t,  s^{t+1}) = 
    \left\{
    \begin{aligned}
         &\ 0,\text{ $\goal^t > \Gamma(s^{t+1})$ or $a^t$ is  not ``stay''} \\
         &\ \frac{|\goal^t-\Gamma(s^{t+1})|}{\goal^t}, \text{otherwise} \\
        \end{aligned}
    \right.
\end{equation}
where $\Gamma(s)$ is a pre-defined function that extracts the duration of the agent spent in the current location from state $s$, $g^t$ is the temporal constraint from system dynamics at time $t$. Intuitively, when $\Gamma(s^{t+1})\leq g^t$, $r_J$ should be positive if the agent takes $a^t$ of staying in the current location, indicating the agents should stay in the location to meet the constraint $g^t$. With such a design, the agent's action will be rewarded when it meets the system constraints, and the generated movement is more likely to be a real-world movement.  

The final surrogate reward for training $\Policy_\theta$ is defined as follows:
\begin{equation}
\label{eq:gail-reward-all}
    r = (1-\eta)\cdot r_J(s^t, a^t,\goal^t,  s^{t+1}) + \eta\cdot  r_D(s^t, a^t, \goal^t)
\end{equation}
where $\eta$ is a hyper-parameter that balances the objective of satisfying the physical constraints and mimicking the true trajectories, both of which are pushing the policy learning towards modeling real-world transitions. 

\subsection{Training Process}
\label{sec:method-discussions}
The training procedure of \ours is an iterative process of learning policy $\Policy_\theta$, discriminator $\DiscT$ and system dynamics $\System_\delta$. We firstly initialize the parameters of $\System_\delta$, $\Policy_\theta$ and $\DiscT$ and pre-train $\System_\delta$. At each iteration of the algorithm, the policy parameters are used by every agent to generate trajectories $\Traj_G$. Rewards are then assigned to each state-action-goal pair in these generated trajectories. Then the generated trajectories are used to perform an update on the policy parameters $\theta_i$ via TRPO~\cite{Schulman2015Trust} in reinforcement learning. Generated trajectories $\Traj_G$ and observed true trajectories $\Traj_E$ are subsequently used as the training data to optimize parameters $\psi$. Specifically, in learning $\System_\delta$, we use the observed trajectories of all agents and build training data for system dynamics $\System_\delta$ by extracting features $o_g$ for a location and calculating the duration that each agent spent in the location as labels $\goal$. In this paper, we use the location ID, time, the number of agents in the location as features, and predict constraints $\goal$ as the duration an agent stays in the location.

\section{Experiment}

\subsection{Experimental Environments and Datasets}

\begin{figure}[t!]
  \centering
  \includegraphics[width=.9\linewidth]{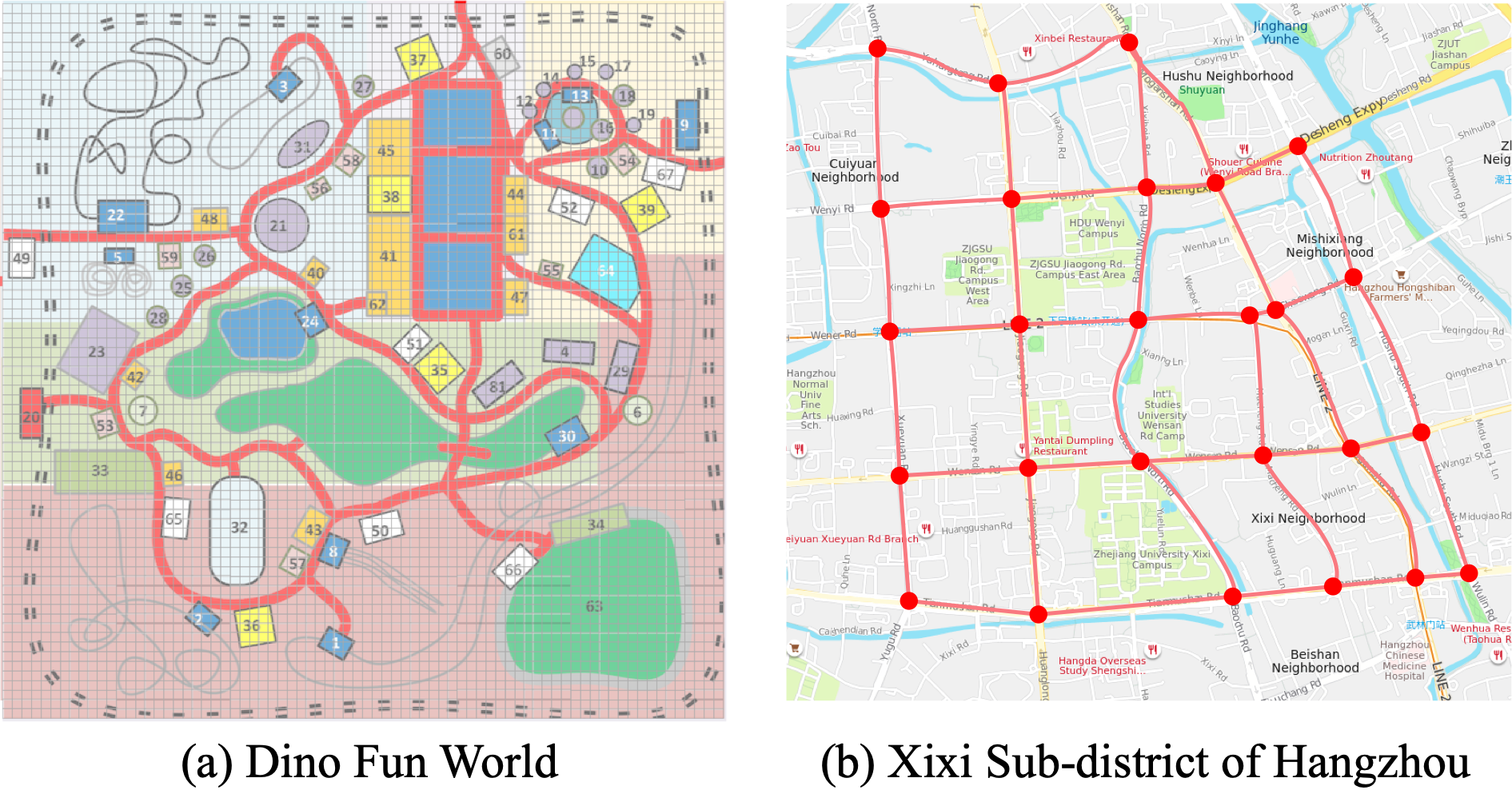}
  \caption{Experiment environments in this paper. Left: Map of Dino Fun World in \themepark. Right: Xixi Sub-district of Hangzhou, China in \routecity.}
  \label{fig:data}
  \vspace{-3mm}
\end{figure}

We evaluate our method in two real-world travel datasets: the travel behavior data in a theme park and the travel behavior of vehicles in a road network. The state and action definitions in each environment are shown in Table~\ref{tab:data-park}. The detailed statistics of the dataset can be found in the Appendix.
~\noindent\\$\bullet$ \textbf{\themepark}. This is an open-accessed dataset\footnote{http://vacommunity.org/VAST+Challenge+2015} that contains the tracking information for all visitors to a theme park, DinoFun World, as is shown in Figure~\ref{fig:data} (a).
DinoFun World covers a geographic space of approximately 500x500 $m^2$ with ride attractions, and hosting thousands of visitors each day. All visitors must use a mobile application which records the location of visitors by a grid ID, where the whole park is divided into $5m\times 5m$ grid cells. 
Each data record contains a record time, a traveler ID, a grid ID, and an action. The action is recorded every second when travelers move from grid square to grid square or check-in at attractions. 
~\noindent\\$\bullet$ \textbf{\routecity}. This is a vehicle trajectory dataset captured from surveillance cameras installed around 25 intersections in Xixi Sub-district at Hangzhou, China, as is shown in Figure~\ref{fig:data} (b). The trajectory of a recorded vehicle includes the timestamp, vehicle ID, road segment ID, and the action of the vehicle. The action can be staying on the current road segment, and transiting to the next road segment by turning left/right, taking U-turn, or going straight.

\begin{table}[t!]
\small
\caption{State and action definition for travelers/vehicles in \themepark and \routecity. In \themepark, travelers move between grids; the actions indicate which neighboring grid it travels or stay/check-in in the current grid. In \routecity, vehicles move between road segments, and the actions indicate to stay in the current road or to go to different directions towards its neighboring roads.}
\label{tab:data-park}
\begin{tabular}{lll}
\toprule
Env & Type & Description \\ \midrule
\parbox[t]{2mm}{\multirow{5}{*}{\rotatebox[origin=c]{90}{\themepark}}} 
 & Time     & \begin{tabular}[c]{@{}l@{}}Time spent in current grid, start time\end{tabular}             \\
 & Location & \begin{tabular}[c]{@{}l@{}}
 Current grid ID, population in current grid,\\
 if current grid checkinable\\
 \end{tabular}\\
\cline{2-3} 
 & Action & \begin{tabular}[c]{@{}l@{}} Move to eight adjacent grids, \\ stay or check-in in current grid \end{tabular}  \\
\midrule
\parbox[t]{2mm}{\multirow{6}{*}{\rotatebox[origin=c]{90}{\routecity}}}
 & Time     & \begin{tabular}[c]{@{}l@{}}Time spent in current road, start time\end{tabular}             \\
 & Location &\begin{tabular}[c]{@{}l@{}}
 Current road ID, destinated road, \\
 current population in the grid, previous road
 \end{tabular} \\
 & Activity & \begin{tabular}[c]{@{}l@{}}Last action \end{tabular} \\ \cline{2-3} 
 &  Action & \begin{tabular}[c]{@{}l@{}}Turn left, turn right, take u-turn, \\ go straight, stay on current road \end{tabular}  \\
\bottomrule
\end{tabular}
\vspace{-3mm}
\end{table}

\begin{figure*}[t]
  \centering
  \includegraphics[width=0.95\textwidth]{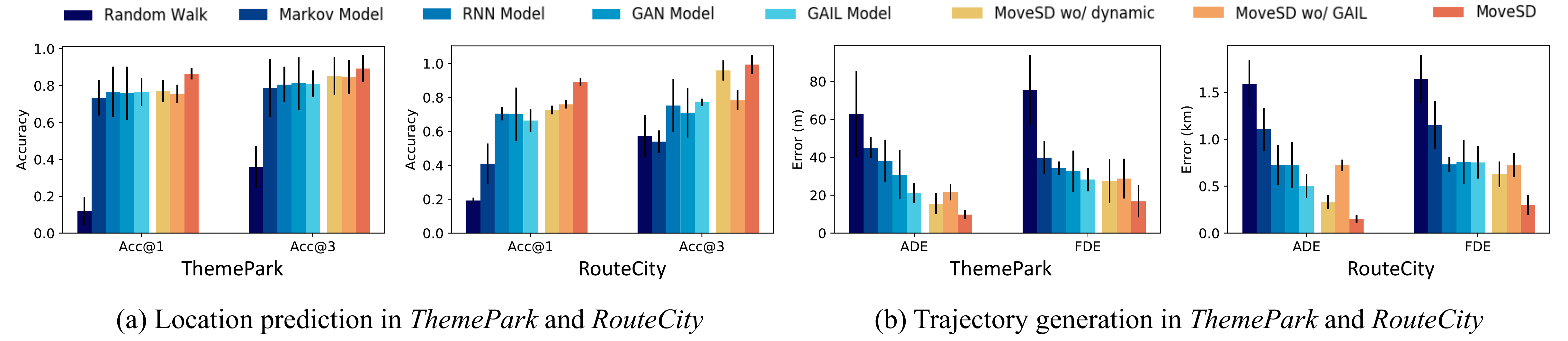}
  \caption{Performance of different methods, including state-of-the-art baselines and the variants of our proposed model. (a) The accuracy (Acc@N) of predicting the next location. The higher, the better. (b) The average displacement error (ADE) and final displacement error (FDE) of generating future trajectories. The lower, the better. \ours performs the best against state-of-the-art baselines and its variants. Best viewed in color.}
  \label{fig:overall}
  \vspace{-3mm}
\end{figure*}

\subsection{Baselines}
We compare with both classical and state-of-the-art methods in human mobility prediction and imitation learning algorithms. We use the same features in deep learning-based methods and our proposed method for a fair comparison.
~\noindent\\$\bullet$ \textbf{\random}~\cite{brockmann2006scaling}. This is a classic method that models the movement of agents as a stochastic process, where the agent takes action among all possible actions with equal probability.
~\noindent\\$\bullet$ \textbf{\markov}~\cite{gambs2012next}.  The Markov-based method regards all the visited locations as states and builds a transition matrix to capture the first or higher-order transition probabilities between them. Following existing methods~\cite{feng2018deepmove,gao2019predicting}, we use the first-order Markov model in our experiment.
~\noindent\\$\bullet$  \textbf{\rnn} is widely used to predict the next location by modeling temporal and spatial history movements~\cite{liu2016predicting,gao2019predicting,feng2018deepmove}. In this paper, we use LSTM in~\cite{liu2016predicting} as the base network of our proposed model and as a baseline. 
~\noindent\\$\bullet$ \textbf{\gan}~\cite{sadeghian2019sophie} uses the framework of generative adversarial networks to generate the next location based on past states without explicitly modeling the agent's actions~\cite{gupta2018social,sadeghian2019sophie,ma2019trafficpredict,feng2020learning}.
~\noindent\\$\bullet$ \textbf{\gail} considers the actions of the agents explicitly by learning the decision policy with generative adversarial networks~\cite{ho2016generative,song2018multi,zheng2020learning,wei2020imingail}. Different from our proposed model, \gail uses feed-forward networks without the system dynamics.

\subsection{Tasks and Metrics}
To measure the discrepancy between the learned state transition model and real-world transitions, we evaluate the performance of different methods in the following two tasks:

\textbf{Next location prediction.} Given the same input states for an agent, a good state transition model should perform well in predicting the next location. Following existing studies~\cite{song2010limits, song2016prediction,feng2018deepmove}, we use the standard evaluation performance metrics, \textbf{Acc@k}, which ranks the candidate next locations by the probabilities generated from the model, and check whether the ground-truth location appears in the top $k$ candidate locations. 

\textbf{Trajectory generation.}
Given the same initial states for an agent, a good state transition model should not only perform well in predicting the next location but also be precise in generating the future trajectories as observed ones. Therefore, we also evaluate the performance on generating trajectory to measure the discrepancy between the learned transition model and real transitions, with the following metrics widely used in existing literature~\cite{zou2018understanding,lisotto2019social,liang2019peeking}:
~~\noindent\\$\bullet$ \textbf{Average Displacement Error (ADE)}: The average of the root mean squared error (RMSE) between the ground truth coordinates $Y_i^t$ and the predicted coordinates:$\hat{Y}_i^t$ at every timestamp $t$ for every trajectory $i$: 
$
ADE = \frac{\sum_{i=1}^N\sum_{t=1}^{T_{pred}}\sqrt{(\hat{Y}^t_i-Y^t_i)^2}}{N\cdot T_{pred}}
$
~~\noindent\\$\bullet$ \textbf{Final Displacement Error (FDE)}: The average of the RMSE at the final predicted points $\hat{Y_i}$ and the final true points $Y_i$ of all trajectories:
$
FDE = \frac{\sum_{i=1}^N\sqrt{(\hat{Y}_i-Y_i)^2}}{N}
$


\subsection{Experimental Settings}
We specify some of the important parameters here and all the parameter settings can be found in our codes.
In all the following experiments, if not specified, the observed time length $L_{in}$ is set to be 10. The output length of $L_{out}$ is 1 for the next location prediction task, and 1000 for trajectory generation task. We fix the length $L_{in}$ and $L_{out}$ for simplicity, but our methods can be easily extended to different lengths since the neural networks are recurrent in taking the trajectories as input and in predicting future trajectories.  We sample the trajectories at every second for \themepark, and at every 10 seconds for \routecity. $\eta$ is set as 0.8. 

\subsection{Overall Performance}
In this section, we investigate the performance of our proposed method \ours on learning the travel movements in two real-world datasets from two perspectives: predicting the next location and generating future trajectories. Figure~\ref{fig:data} shows the performance of the proposed \ours, classic models as well as state-of-the-art learning methods in the real-world environments. 
We have the following observations:
~\noindent\\$\bullet$ \ours achieves consistent performance improvements over state-of-the-art prediction and generation methods (\rnn and \gan respectively) across different environments. The performance improvements are attributed to the benefits from the modeling of the decision process with \gail and the integration with system dynamics. We also noticed that \gail achieves better performance than \gan in most cases. This is because \gail explicitly model the movement policy of agents, which conveys the importance of modeling the decision-making process in human movement. 
~\noindent\\$\bullet$ The performance gap between the proposed \ours and baseline methods becomes larger in trajectory generation than in the next location prediction. This is because the framework of \gail, which iteratively updates policy $\pi_\theta$ and discriminator $\DiscT$ enables us to model the decision process that determines the whole trajectory effectively. The discriminator in our model can differentiate whether the trajectory is generated or real, which can drive the bad-behaved actions in one iteration to well-behaved ones during the policy learning process of the next iteration.


\subsection{Ablation study}
\label{exp:system-dynamic}

In this section, we perform an ablation study to measure the importance of learning the decision-making policy and the system dynamics.
~\noindent\\$\bullet$ \textbf{\ourswogail}. This model takes both the state and constraints from system dynamics as its input and output the next location directly, which does not learn the movement policy in the decision-making process of the agent. It can also be seen as \rnn methods with the information of system dynamics. 
~\noindent\\$\bullet$ \textbf{\ourswo}. This method uses \gail but does not consider the system dynamics. That is, the input of the policy and discriminator do not contain $\goal$ from system dynamic model $\System$.
~\noindent\\$\bullet$ \textbf{\ours}. This is our final method, which uses the adversarial learning process as \gail, and considers the system dynamic. 

Figure~\ref{fig:overall} shows the performance of variants of our method.
We have the following observations:
~\noindent\\$\bullet$ \ours outperforms both \ourswo and \ourswogail, indicating the effectiveness of using \gail and the system dynamics. Compared with \ourswogail, \ours performs better because it learns the decision-making process as the underlying mechanism behind the movements. Compared with \ourswo, \ours takes the system constraint $\goal$ into consideration, learning movement policies to avoid those infeasible actions that do not comply with the system dynamics.
~\noindent\\$\bullet$ We also notice that \ourswogail outperforms pure \rnn in most cases. This is because \ourswogail has extra information on the constraints from system dynamics within the model, which validates the effectiveness of considering system dynamics.
In the rest of our experiments, we only compare \ours with other methods.

To get deeper insights on \ours, we also investigated the following aspects about our model and their empirical results are left in the appendix due to page limits: (1) different paradigms in training $\System$, (2) the necessity of stochastic modeling of $\System$ and (3) sensitivity study. We also provided a case study for better understanding.


\section{Conclusion}

In this paper, we present a novel perspective in modeling the state transition of human movement in space as a decision process and learns the decision making policy with system dynamics. Specifically, we argue it is important to integrate the constraints from the system dynamic to help the learned state transition more realistic.  Extensive experiments on real-world data demonstrate that our method can not only accurately predict the next state of agents, but also accurately generate longer-term future movements.

While \ours substantially provides insights to model human movements from decision making perspective with system constraints, we believe that this problem merits further study. One limitation of our current method is that, we model the human movements over discrete action space, whereas the movement of human over free space is continuous. An exciting direction for future work would be to develop stronger learning models for continuous action space. Second, the raw data for observation only include the status of travelers or vehicles and the location information. More exterior data like weather conditions might help to boost model performance. Lastly, we can further model different modalities in policies for different kinds of agents like private cars and taxis.
\section{Acknowledgments}
The work was supported in part by NSF awards \#1652525 and \#1618448. The views and conclusions contained in this paper are those of the authors and should not be interpreted as representing any funding agencies.

\bibliography{sample-base}

\newpage
\newpage
\section{Appendix}
\label{sec:addendum}

\subsection{A. Dataset Description}
\label{sec:add:dataset}
In this paper, we use two real-world human movement datasets:

\themepark: This is an open-accessed dataset that contains the tracking information for all of the paying park visitors to a simulated park, DinoFun World. DinoFun World covers a large geographic space (approx. 500x500 $m^2$) and is populated with ride attractions, restaurants, and food stops, and hosting thousands of visitors each day. All visitors to the park use a parking app to check into rides and some other attractions. The park is equipped with sensor beacons that record movements within the park. Sensors are sensitive within a $5m\times 5m$ grid cell. Each data record contains a record time, a traveler ID, a grid ID with coordinate locations, and an action. The action is recorded every second when travelers move from grid square to grid square or “check-in” at rides, meaning they either get in line or onto the ride.

\routecity. This is a trajectory dataset of vehicles captured from surveillance cameras installed around 25 intersections in Xixi Sub-district at Hangzhou, China. Each data record consists of a record time, a vehicle ID, an intersection ID. By analyzing these records with camera locations, the trajectory of a vehicle can be recovered as a timestamp, a vehicle ID, a road segment ID, and an action of the vehicle. The actions can be inferred from the consecutive road segments at every 10 seconds. The vehicles have actions like keeping traveling on the current road segment or transiting to the next road segment by turning left, turning right, taking U-turn, or going straight.

\subsection{B. Model Parameters and Configurations}
\label{sec:add:parameters}
The hyper-parameters for our network structure and training process is shown in Table~\ref{tab:parameters}.

\begin{table*}[t!]
\centering
\caption{Parameter settings}
\label{tab:parameters}
\small
\begin{tabular}{p{6.5cm}c|p{6.5cm}c}
\toprule
Parameter                                            & Value   & Parameter                                                       & Value     \\ \midrule
Number of hidden layers $G$ in $\System$ & 2       & Feature length $n$ of hidden layers in $\System$   & 50        \\
Input feature dimension $c$                          & 3       & Output dimension                                                & 2         \\
Number of recurrent units                      & 10      &  Embedding size $m_1$                              & 50        \\
Number of hidden layers $H$ in Recurrent module    & 2       & Feature length $d$ of hidden layers for MLPs              & 50        \\
Activation function of hidden layers                & ReLU    & Activation function of output layer in $\Policy$                               & Softmax   \\
Activation function of output layer in$\System$                    & ReLU    & Activation function of output layer in $\Disc$    & Sigmoid                                                \\ \midrule
\multicolumn{2}{c}{Vocabulary size in embedding $l$} & \multicolumn{2}{c}{10000 in \themepark, 78 in \routecity} \\
\multicolumn{2}{c }{State feature dimension $k$ } & \multicolumn{2}{c}{6 in \themepark, 12 in \routecity}\\
\multicolumn{2}{c}{Number of actions $|\mathcal{A}|$ }    & \multicolumn{2}{c}{11 in $\themepark$, 5 in \routecity}  \\
\multicolumn{2}{c}{Total training iterations for generator and discriminator update } & \multicolumn{2}{c}{100} \\ \midrule
Batch size                & 128    & Learning rate    & 0.0003   \\
Discount factor $\gamma$ for TRPO & 0.8    &  Gradient clip threshold for TRPO    & 0.2 \\  
$\lambda$ of Advantage Estimator for TRPO & 0.98    &  Maximum iterations for GAIL    & 100 \\
$\Disc$ updating frequency per iteration & 5    &  $\Policy$ updating frequency per iteration  & 100 \\
\bottomrule
\end{tabular}
\end{table*}

\subsection{C. Model update for Beta Distribution in $\System$}
\label{sec:add:beta}

When the targets in $\System$ are real-valued scalars with lower and upper bounds, we can shift and rescale the values to be in the range [0, 1] and model the target as a sample from a Beta distribution. It can be parameterized by a neural network that outputs two values $\alpha, \beta > 0$.
The relationships between the temporal constraint $\goal$ and the shaping parameters $\alpha, \beta$ are as follows:
\begin{equation}
    p_{\alpha,\beta}(\goal) = \goal^{\alpha-1}(1-\goal)^{\beta-1}\frac{\Gamma(\alpha\beta)}{\Gamma(\alpha)\Gamma(\beta)}
\end{equation}
with log-likelihood:
\begin{equation}
\begin{aligned}
    ln\ p_{\alpha,\beta}(\goal) & = (\alpha-1)ln(\goal)+(\beta-1)ln(1-\goal) \\
    & + ln\ \Gamma(\alpha\beta) - ln\ \Gamma(\alpha) - ln\ \Gamma(\beta)
\end{aligned}
\end{equation}
and gradients:
\begin{equation}
\label{eq:beta}
\begin{aligned}
    \frac{\partial}{\partial \alpha} ln(p_{\alpha,\beta}(\goal)) = ln(\goal) +  \Psi(\alpha\beta) - \Psi(\alpha) \\
    \frac{\partial}{\partial \beta} ln(p_{\alpha,\beta}(\goal)) = ln(1-\goal) + \Psi(\alpha\beta) - \Psi(\beta)
\end{aligned}
\end{equation}
where $\Psi$ is the digamma function $\Psi(x)=\frac{\partial}{\partial x}ln(\Gamma(x))$.

\begin{algorithm}[!ht]
\DontPrintSemicolon
\caption{Training procedure of \ours with options}
\label{alg:learntosim1}

\KwIn{Observed true trajectories $\Traj_E$, initial system dynamics, policy and discriminator parameters $\delta_0$, $\theta_0$, $\psi_0$, training option}
\KwOut{Policy $\Policy_\theta$, Discriminator $\DiscT$, System dynamics $\System_\delta$}

\If{Option 1}{
$\bullet$ Pretrain system dynamics $\System_\delta$ using Eq.~\eqref{eq:beta}\;} 
\For{i $\longleftarrow$ 0, 1, \dots}
{
    $\bullet$ Rollout trajectories for all agents 
    $\Traj_G = \{\traj |\traj =( \traj^{t_0},\cdots ,\traj^{t_T} )\}$, where $\traj^{t}=(s^{t}, a^{t}, g^{t}, s^{t+1})$, $g^{t}= \System_\delta(o^t_g)$, and
    $a^{t}\sim \Policy_{\theta_i}(s^{t}, a^{t}, g^{t})$; \;
    
    \# Generator update step \;
    
    $\bullet$ Score $\traj^{t}$ from $\TrajD_G$  with discriminator, generating reward using Eq.~\eqref{eq:gail-reward}; \;
    
    \If{Option 2}{
    $\bullet$ Update system dynamics $\System_\delta$; \;
    $\bullet$ Update $\theta$ given $\TrajD_G$ by optimizing Eq.~\eqref{eq:gail}; \;
    }\ElseIf{Option 3}{
    $\bullet$ Update $\theta$ and $\delta$ given $\TrajD_G$ by optimizing Eq.~\eqref{eq:gail} and using Eq.~\eqref{eq:beta}; \;
    }
    \Else{
    $\bullet$ Update $\theta$ given $\TrajD_G$ by optimizing Eq.~\eqref{eq:gail}; \;
    }
    \# Discriminator update step; \;
    $\bullet$ Update  $\psi$ by optimizing Eq.~\eqref{eq:loss}; \;
}
\end{algorithm}

\subsection{D. Detailed algorithms for different options in  training paradigms}
\label{sec:add:detail}
\begin{figure*}[t!]
  \centering
  \begin{tabular}{cc}
  \includegraphics[width=.45\linewidth]{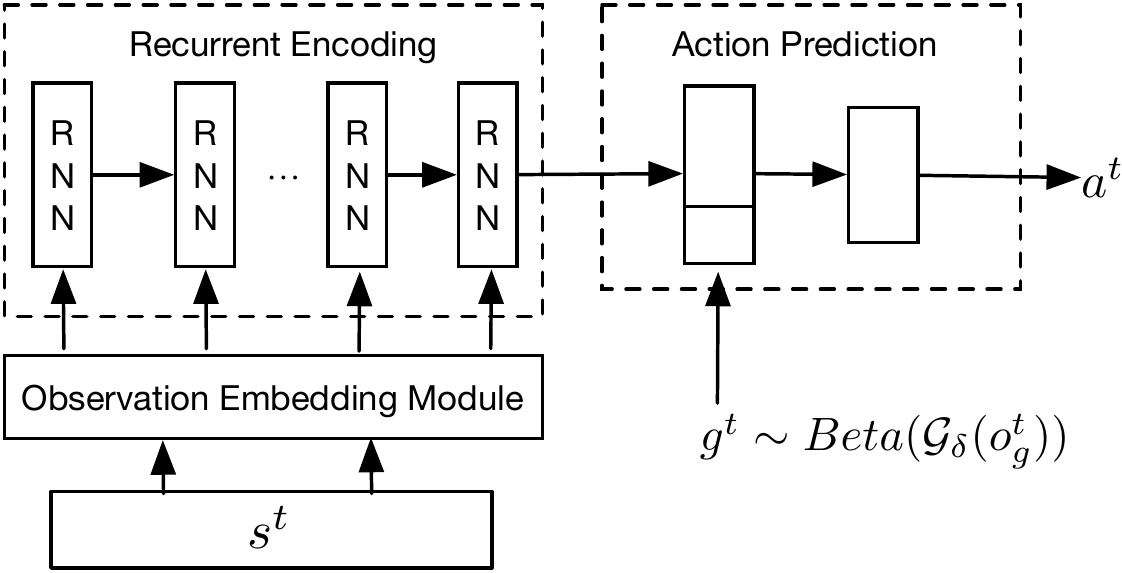} &
  \includegraphics[width=.45\linewidth]{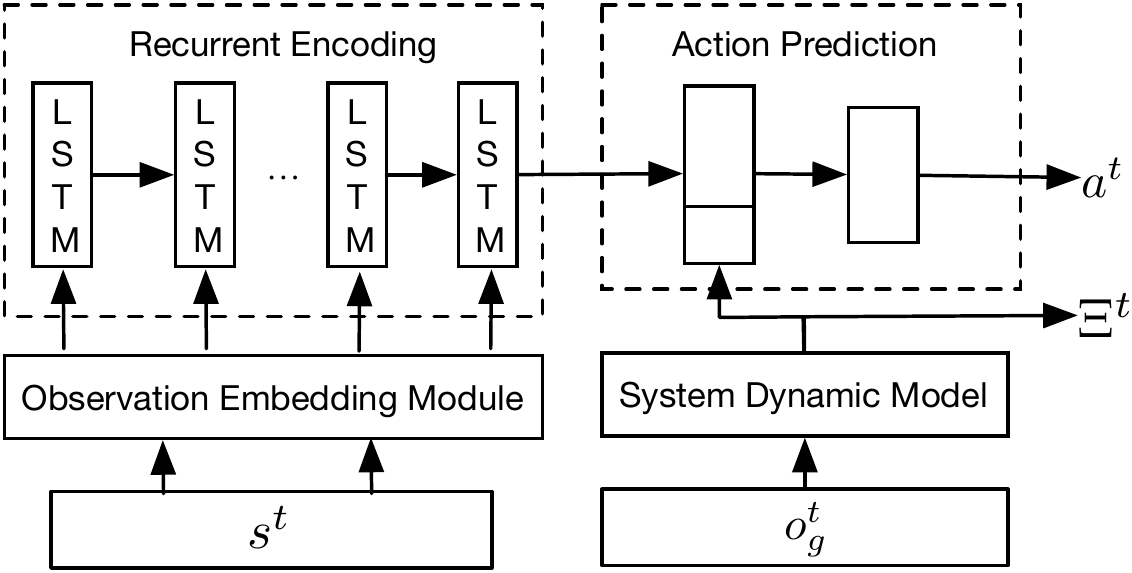} \\
  (a) Option 1 \& 2 & (b) Option 3\\
  \end{tabular}
  \caption{Different architectures of policy network}
  \label{fig:policynetwork3}
\end{figure*}

Since the system constraint $\goal$ given by $\System_\delta$ is used in $\traj$ when learning $\Policy_\theta$ and $\DiscT$, here we discuss the options in learning $\System_\delta$, and investigate their effectiveness empirically in Appendix D-2.
~\noindent\\$\bullet$ \textbf{Option 1 (Default)}: Pre-train $\System_\delta$ before the learning of $\Policy_\theta$ and $\DiscT$ (line 1 in Algorithm~\ref{alg:learntosim1}), fix $\delta$ during the learning of $\theta$ and $\psi$. The architecture for this network is shown in Figure~\ref{fig:policynetwork3}
~\noindent\\$\bullet$ \textbf{Option 2}: At each round of updating $\Policy_\theta$ and $\DiscT$, we update $\System_\delta$ as well based on the generated trajectories.
~\noindent\\$\bullet$ \textbf{Option 3}: Integrate the $\System_\delta$ with $\Policy_\theta$ in one model, as is shown in Figure~\ref{fig:policynetwork3}(b). The output of $\System_\delta$ is fed into $\Policy$ before action prediction module. The training of this option would require a model that is optimized through multi-task learning, with one task learns the action prediction and the other task learns to predict $\goal$. 

Their training algorithm is shown in Algorithm~\ref{alg:learntosim1}.

\subsection{E. Additional Experiments}

\subsubsection{1. Stochastic/Deterministic modeling of $\System$}
\label{sec:exp-stochastic}
As mentioned in Section~\ref{sec:method:dynamic}, we model the system constraint from $\System$ by modeling the distribution to mimic the stochastic nature of system dynamics. In this section, we investigate whether this stochastic modeling is necessarily by comparing it with deterministic modeling, which directly output a scalar value. 

As is shown in Table~\ref{exp:trainingparadigm}, \ours with stochastic $\System$ achieves consistently better performance than deterministic $\System$. The reason is that there are unobserved factors and stochastic $\System$ learns the distribution of data with probability, which can be used to learn with this kind of missing data. In the rest of the experiments, we only use stochastic $\System$ in \ours and compare it with other baseline methods.

\subsubsection{2. Different training paradigms of $\System$}
\label{sec:exp-paradigm}
As mentioned in Section~\ref{sec:method-discussions}, $\Policy$ can be trained in different ways with regards to $\System$: 1) \emph{pre-trained} $\System$, 2) $\System$ \emph{iteratively trained} with $\Policy$, or 3) $\System$ \emph{jointly learned} with $\Policy$ in one model. 

The results in Table~\ref{exp:trainingparadigm} show that \ours with pre-training $\System$ achieves similar performance with using one model integrating both $\System$ and $\Policy$, but outperforms training $\System$ and $\Policy$ iteratively. The reason could be that if the errors for predicting $\goal$ form $\System$ is large in the current iteration and if the errors cannot propagate back to the learning of $\Policy$, $\Policy$ taking $\goal$ from $\System$ would lead to further error. Therefore, a good start of $\System$ through pre-training or using one model to learn both $\System$ and $\Policy$ can lead to relatively better performance. In the rest of the experiments, we only use pre-trained $\System$ in \ours and compare it with other baseline methods.

\begin{figure*}[t!]
  \centering
  \begin{tabular}{cccc}
  \includegraphics[width=0.22\textwidth]{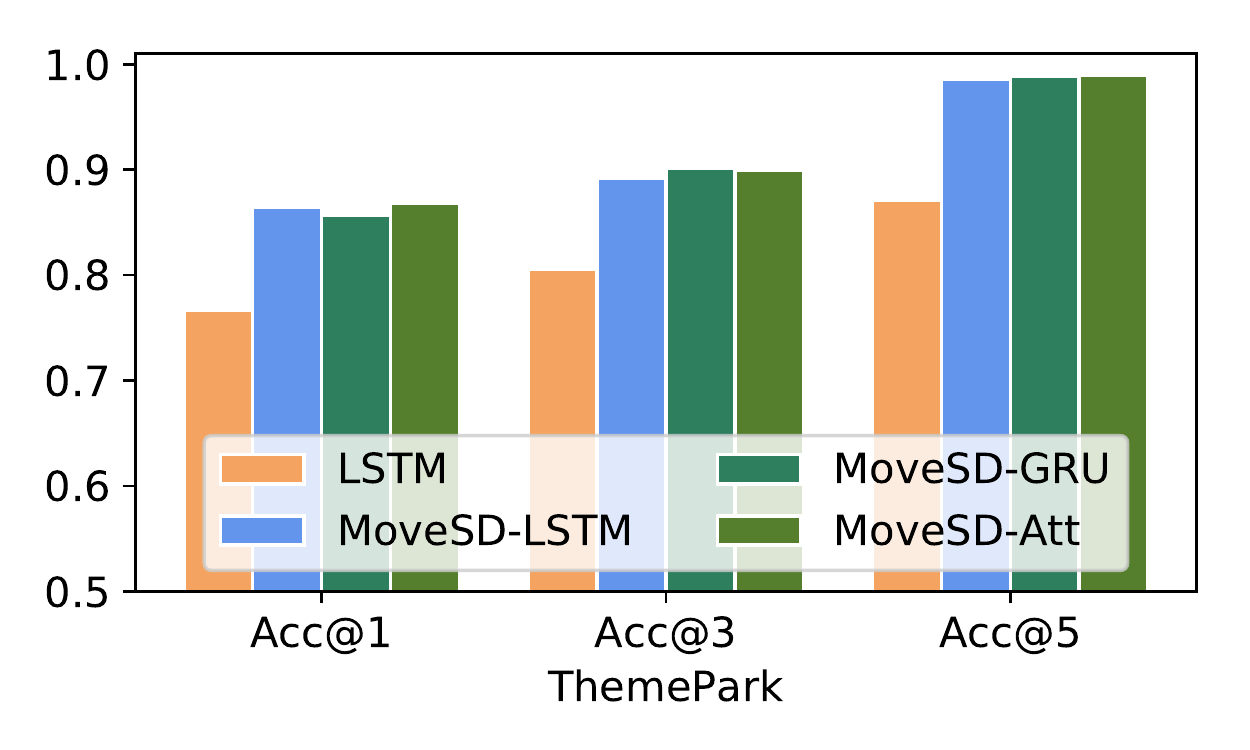}&
   \includegraphics[width=0.22\textwidth]{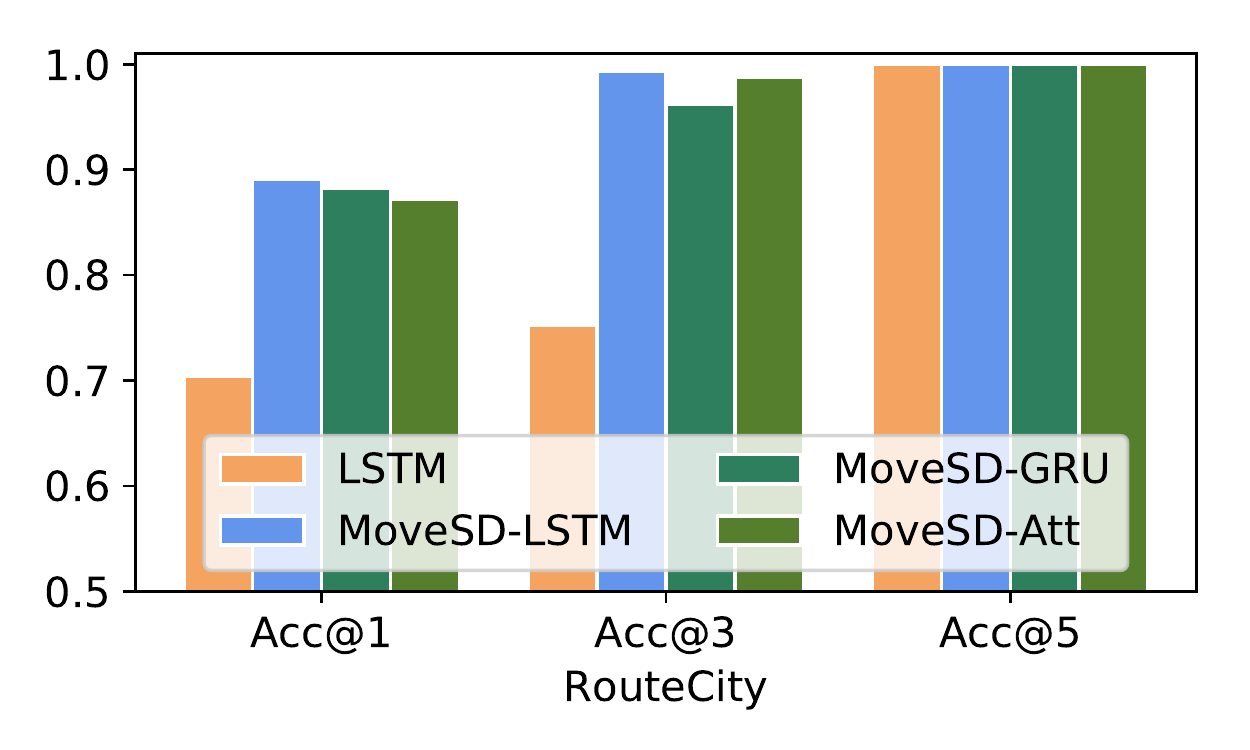} &
   \includegraphics[width=0.22\textwidth]{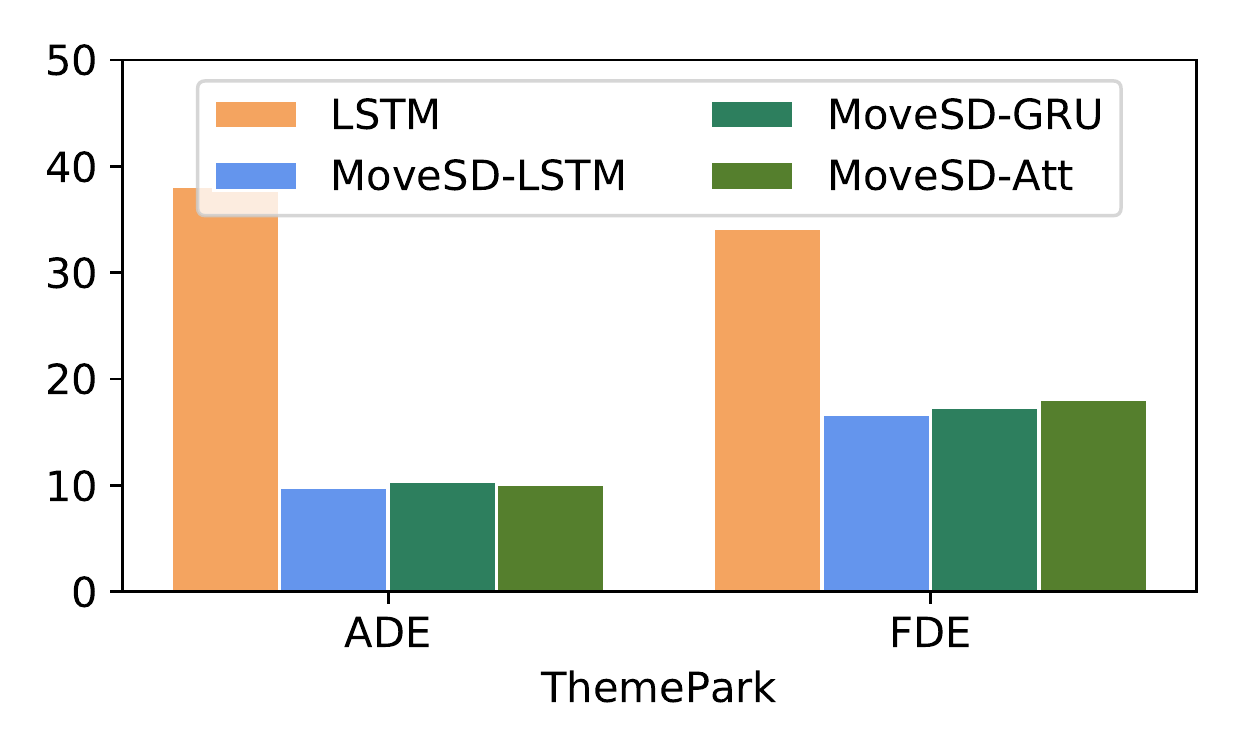}  &
  \includegraphics[width=0.22\textwidth]{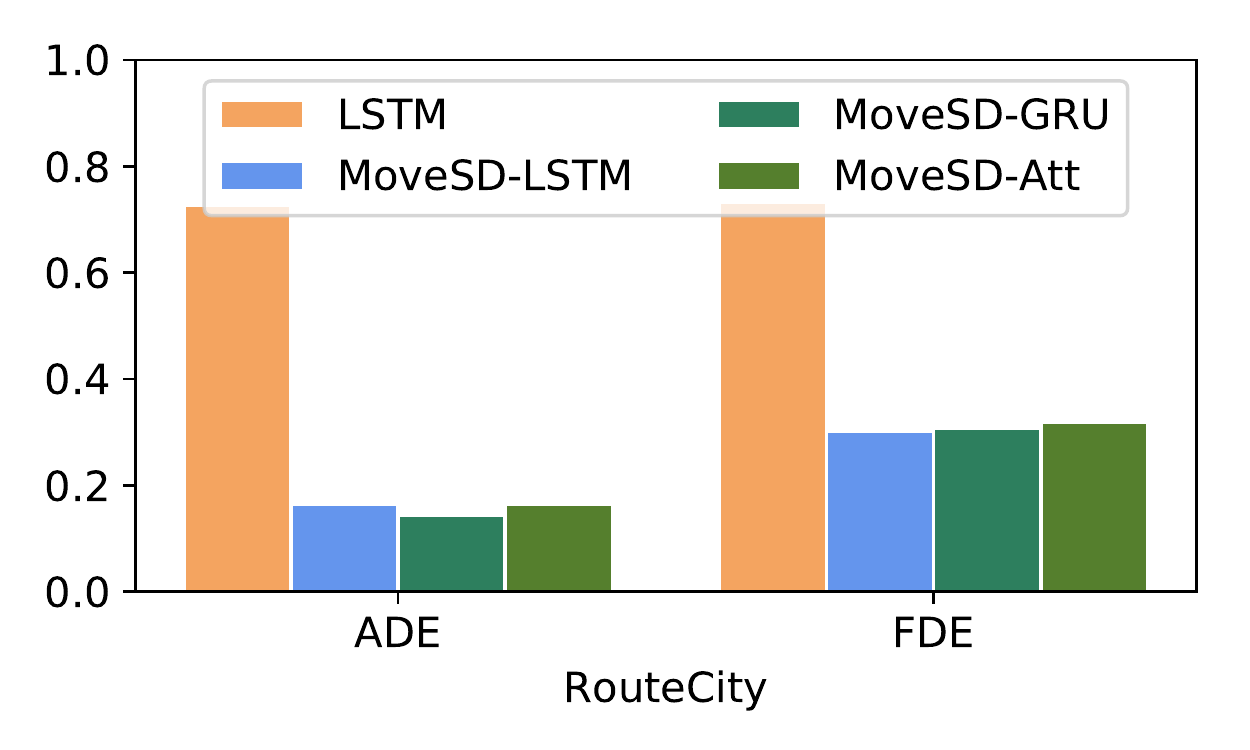}\\
   \multicolumn{2}{c}{(a) Accuracy of location prediction in \themepark and \routecity} &

  \multicolumn{2}{c}{(b) Error of trajectory generation in \themepark and \routecity}  \\
  \end{tabular}
  \caption{Performance of \ours under with different base models, compared with \rnn (LSTM). (a) Accuracy of next location prediction under both scenarios. The higher, the better. (b) Error of trajectory generation under both scenarios. The lower, the better.
  \ours consistently outperforms \rnn. Best viewed in color.}
  \label{fig:base-study}
\end{figure*}

\begin{figure}[t]
  \centering
  \begin{tabular}{cccc}
  \includegraphics[width=0.22\textwidth]{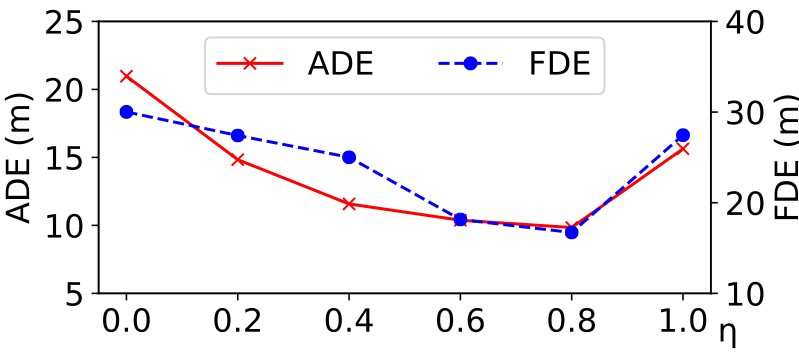}&
   \includegraphics[width=0.22\textwidth]{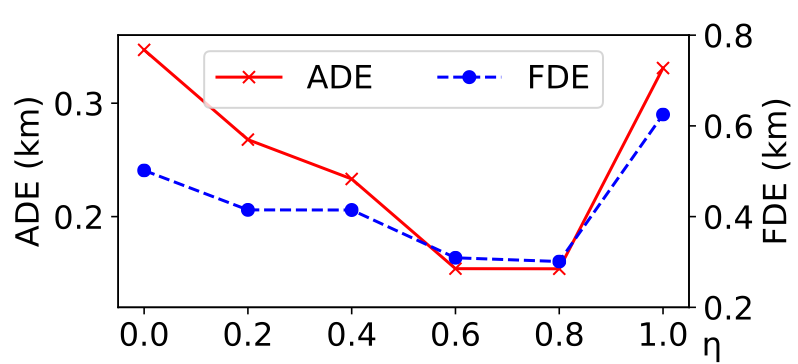} \\
   (a) \themepark & (b) \routecity \\
  \end{tabular}
  \caption{Performance of \ours for trajectory generation with different values of $\eta$. The lower, the better.}
  \label{fig:eta-study}
\end{figure}

\begin{figure*}[b!]
  \centering
  \includegraphics[width=.95\linewidth]{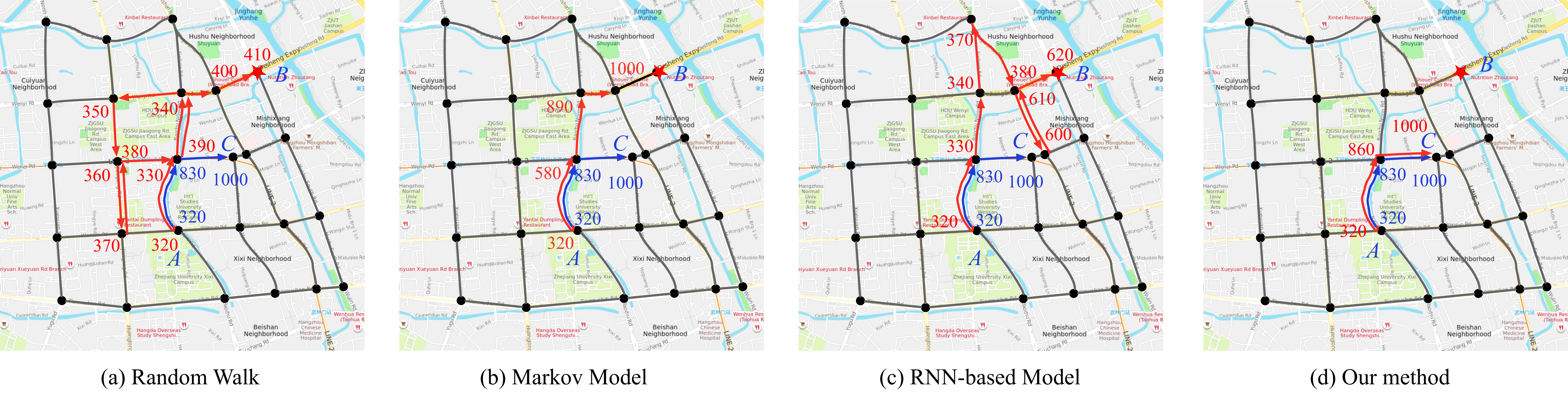}
  \caption{A case study on a vehicle's movement in the first 1000 seconds. All the methods take the same initial information that the vehicle starts its movement at location $A$ at the 320 second, and its destination is location $B$ (shown as the red star). The numbers next to the nodes are the time of vehicles arriving at corresponding intersections. The blue lines are the true observed trajectories of the vehicle, where it only reaches $C$ at the 1000 second.
  The red lines in different figures indicate the trajectories generated by different methods. Only \ours can generate a similar trajectory as observed. Best viewed in color.}
  \label{fig:case}
  \vspace{-3mm}
\end{figure*}

\begin{table}[t!]
\small
\caption{Investigation of stochastic/deterministic modeling of $\System$ and different training paradigms for $\Policy$ and $\System$ in generating trajectories. The lower, the better. Stochastic modeling is better than deterministic modeling. Pre-training $\System$ and jointly learning $\System$ with $\Policy$ achieves slightly better performance than iteratively updating $\System$ and $\Policy$.}
\vspace{-3mm}
\label{exp:trainingparadigm}
\begin{tabular}{p{0.3cm}cp{0.7cm}p{0.8cm}|p{0.8cm}p{0.8cm}}
\toprule
\multirow{2}{*}{}              & \multirow{2}{*}{Options} & \multicolumn{2}{c|}{\themepark} & \multicolumn{2}{c}{\routecity} \\
                               &    & ADE(m)  & FDE(m) & ADE(km) & FDE(km) \\ \midrule
\multirow{3}{*}{\rotatebox[origin=c]{90}{\begin{tabular}[c]{@{}c@{}}Deter-\\ministic\end{tabular}}} & \begin{tabular}[c]{@{}l@{}}\#1: 
Pre-trained $\System$
\end{tabular} 
& 15.653        & 22.231        & 0.236         & 0.498         \\
   & \begin{tabular}[c]{@{}c@{}}\#2: Iterative training\end{tabular}
                               & 17.829        & 25.932        & 0.270         & 0.531         \\
   & \begin{tabular}[c]{@{}c@{}}\#3: One model\end{tabular}     
                               & 14.325        & 22.617        & 0.242         & 0.515         \\ \midrule
\multirow{3}{*}{\rotatebox[origin=c]{90}{\begin{tabular}[c]{@{}c@{}}Stoch-\\astic\end{tabular}}}      & \begin{tabular}[c]{@{}l@{}}\#1: 
Pre-trained $\System$
\end{tabular}  &  \textbf{9.842}  & \textbf{16.716}  & \textbf{0.154} & 0.301    \\
     & \begin{tabular}[c]{@{}c@{}}\#2: Iterative training\end{tabular} & 13.122        & 20.970        & 0.189         & 0.343         \\
                               & \begin{tabular}[c]{@{}c@{}}\#3: One model
\end{tabular}    & 10.399   & 17.806    & 0.160 & \textbf{0.292}   \\
\bottomrule
\end{tabular}
\vspace{-3mm}
\end{table}

\subsubsection{3. Sensitivity Study: different base models in $\Policy$}
\label{exp:base-model}
In this section, we investigate the impact of base RNN models in $\Policy$ proposed in the existing literature - more specifically, GRU as in~\cite{gao2019predicting}, attentional RNN as in~\cite{feng2018deepmove}, and LSTM~\cite{liu2016predicting}, termed \ours-GRU, \ours-Att, and \ours-LSTM respectively. Figure~\ref{fig:base-study} summarizes the experimental results. We have the following observations:
~\noindent\\$\bullet$ Our proposed method performs steadily under different base models across various tasks, indicating the idea of our method is valid across different base models. In the rest of our experiments, we use LSTM as our base model and compare it with other baseline methods.
~\noindent\\$\bullet$ Our proposed method with different base models performs consistently better than the baseline method $\rnn$. This is because our method not only learns to model the behavior but also considers the influence of system dynamics.

\subsubsection{4. Sensitivity Study: different values of balancing parameter $\eta$}
\label{exp:balancing}
In this section, we study the influence of $\eta$ in Equation~\eqref{eq:gail-reward-all}, which balances the reward from the discriminator $\DiscT$ and the dynamics judger $\judge$. As is shown in Figure~\ref{fig:eta-study}, when we use a combined reward from both $\judge$ and $\DiscT$, \ours performs better than purely from $\judge$ $(\eta=0)$ and purely from $\DiscT$ $(\eta=1)$. In this paper, we set $\eta$ as 0.8 as it performs the best empirically.

\subsection{F. Case Study}
In this section, we showcase the trajectory of a vehicle generated by different methods under \routecity, as is shown in Figure~\ref{fig:case}. Blue lines are the true observed trajectories of the vehicle, where we can see the observed vehicle starts traveling at the 320 seconds, passing through two road segments toward its destination (red star) and only reaches at $C$ at the 1000 second. It spends a certain time on each road, due to the physical length of roads and the travel speed on the road. The red lines indicate the trajectory generated by different methods. We can see that only \ours can generate a similar trajectory as observed, while all other baselines generate unrealistic trajectories. Specifically, since \rnn yields inaccurate predictions when encountering a new state and the generated trajectory drifts away eventually because of the accumulated inaccuracy in previous predictions. On the other hand, \ours explicitly models the mechanism behind state transitions and learns to generate trajectories with physical constraints.

\subsection{G. Detailed results for Figure~\ref{fig:overall}}

Table~\ref{tab:overall} shows the numerical results of Figure~\ref{fig:overall}. \ours performs the best against state-of-the-art baselines and its variants.

\begin{table*}[ht]
\small
\caption{Performance of different methods in predicting the next location and generating future trajectories w.r.t accuracy (Acc@N), average displacement error (ADE), and final displacement error (FDE). For Acc@N, the higher, the better; for ADE and FDE, the lower, the better. \ours performs the best against state-of-the-art baselines and its variants.}
\label{tab:overall}
\begin{tabular}{cccccc|ccccc}
\toprule
            & \multicolumn{5}{c|}{\themepark}                                                          & \multicolumn{5}{c}{\routecity}                                                          \\
            & \multicolumn{3}{c}{Predicting next location} & \multicolumn{2}{c|}{Generating 1000 steps} & \multicolumn{3}{c}{Predicting next location} & \multicolumn{2}{c}{Generating 1000 steps} \\
            & Acc@1  & Acc@3   & Acc@5   & ADE(m)   & FDE(m)   
            & Acc@1  & Acc@3   & Acc@5   & ADE(km)   & FDE(km)   \\ \midrule
\random & 0.120     & 0.358  & 0.574   & 67.291     & 75.492 & 0.192 & 0.573 & 1.000 $^*$  &  1.589  &  1.644              \\
\markov     & 0.734     & 0.788   & 0.795     & 45.087  &  39.768  & 0.408    & 0.539   &  1.000       & 1.105  &  1.149  \\
\rnn    & 0.767     & 0.806  & 0.871   & 38.129   & 34.196 &  0.704    & 0.752   &  1.000     & 0.727 & 0.733 \\ 
\gan  & 0.759    &  0.812  & 0.925   & 30.861    & 32.689  &   0.701   & 0.700  &  1.000     & 0.723  &  0.757 \\
\gail & 0.766    &  0.810  & 0.938   & 20.928    & 28.222  &   0.664   & 0.771  &  1.000     & 0.501  &  0.752 \\
\midrule
\ourswo   & 0.771    &  0.853  & 0.979   & 15.634    & 27.454  &   0.726   & 0.959  &  1.000     & 0.331  &  0.625 \\
\ourswogail &  0.756    &  0.847  &  0.906  & 21.602    & 28.762  &  0.759 & 0.783 & 1.000      & 0.724  &  0.724 \\
\ours & \textbf{0.864}    &  \textbf{0.892}  & \textbf{0.986}  &\textbf{9.842}  & \textbf{16.716}    &  \textbf{0.892}   & \textbf{0.994}  &   1.000    &   \textbf{0.154}  & \textbf{0.301}    \\ \bottomrule
\end{tabular}
\\\footnotesize{$^{*}$Acc@5 in \routecity are all equals to 1 because the maximum number of next candidate locations is 5.}
\vspace{-3mm}
\end{table*}

\end{document}